# Systematic Analysis of Image Generation using GANs

Rohan Akut[1]
Department of Electronics and Telecommunication
MIT College of Engineering
Pune, India
rohan.akut97@gmail.com

Sumukh Marathe[1]
Department of Electronics and Telecommunication
MIT College of Engineering
Pune, India
sumukh.47@gmail.com

Rucha Apte[1]
Department of Electronics and Telecommunication
MIT College of Engineering
Pune, India
ruchaa.apte@gmail.com

Ishan Joshi[1]
Department of Electronics and Telecommunication
MIT College of Engineering
Pune, India
ishanjoshi2697@gmail.com

Dr. Siddhivinayak Kulkarni[2]
Department of Computer Science & Engg
MIT-WPU
Pune, India
siddhivinayak.kulkarni@mitcoe.edu.in

*Abstract*— Generative Adversarial Networks have been crucial in the developments made in unsupervised learning in recent times. Exemplars of image synthesis from text or other images, these networks have shown remarkable improvements over conventional methods in terms of performance. Trained on the adversarial training philosophy, these networks aim to estimate the potential distribution from the real data and then use this as input to generate the synthetic data. Based on this fundamental principle, several frameworks can be generated that are paragon implementations in several real-life applications such as art synthesis, generation of high resolution outputs and synthesis of images from human drawn sketches, to name a few. While theoretically GANs present better results and prove to be an improvement over conventional methods in many factors, the implementation of these frameworks for dedicated applications remains a challenge. This study explores and presents a taxonomy of these frameworks and their use in various image to image synthesis and text to image synthesis applications. The basic GANs as well as a variety of different niche frameworks are critically analyzed. The advantages of GANs for image generation over conventional methods as well their disadvantages amongst other frameworks are presented. The future applications of GANs in industries such as healthcare, art and entertainment are also discussed.

I. INTRODUCTION

Huge research is conducted in the field of Machine Learning. However, most of these advances focused on supervised learning approaches. These approaches focused on feature extraction and prediction. To eliminate the need of feature extraction deep learning algorithms were invented. These algorithms learnt the features on their own and hence proved much more influential than the traditional Machine Learning approaches. However, all of these algorithms were completely data dependent.

Until recently not much research had been done in the Unsupervised Learning Domain. Generative Adversarial Networks (GAN) can be used to generate data from the current available dataset. This reduces the problem of dataset availability and helps in making the dataset more variable.

GANs were first invented by Goodfellow *et al.* [1] in 2014. GAN consists of two networks namely the Discriminator and Generator. The Generator tries to generates synthetic data which resembles the original distribution while the Discriminator tries to classify whether the data is synthetic or original data. Generally, both the Discriminator and Generator are deep neural networks [2][3].

The GANs are trained in such a way that both the Generator and Discriminator try to minimize their losses. This situation is similar to minimax game in Game Theory where the main aim is to achieve Nash equilibrium [4]. The loss function of the GAN is as mentioned in equation (1).

$$\min_G \max_D V(D,G) = E_{x \sim p_{data}(x)}[\log D(x)] + E_{z \sim p_z(z)}[\log(1 - D(G(z)))] \quad (1)$$

Since their introduction GANs have been used for different applications. To name a few, MIDINET architecture [5] is used for music generation, STACKGAN [6] is used for generating image based on textual description (text to image), DiscoGAN [7] and Texture GAN [8] are used for generating images based on an input image, MocoGAN [9] is used for video generation, MR image GAN [10] is used for biomedical engineering. One of the main use cases of GAN is to generate images based on and conditioned by different inputs.

The paper focuses mainly on different types of GANs that are available for image generation. It consists of a detailed study of GANs which take different types of inputs such as text, image, a combination of text and image etc. However, all the GANs mentioned in this paper give image as an output. The paper focuses on the most recent and state of the art GANs that are available and gives a brief summary of the previous GANs that are available to do the similar task.

The paper presents a taxonomic study of GANs. The paper is developed in three sections: (1) The first section gives a brief introduction of the two most generic GANs available for image generation namely: DCGAN, CGAN. Variants of these GANs are used for different applications hence we have described these GANs in the first section. (2) The second section focuses on the GANs which take text as an input and generate image from the text input. (3) The third section describes different GANs which take image/images as an input and generates a new image as an output. The prominent types of GANs are as shown in Fig. 1.

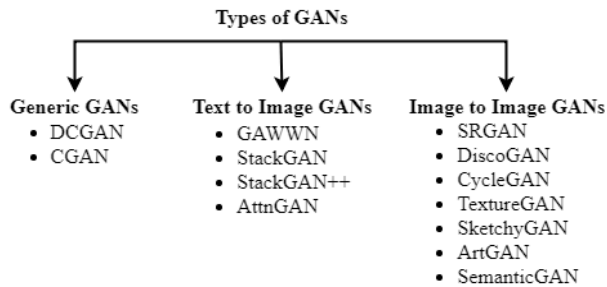

## II. GENERIC GAN

GANs were initially invented by Goodfellow et al.[1] However, the GANs were not generating a high-quality photo realistic image as the images produced by GAN were noisy and not clear. To tackle this, LAPGAN [2] was introduced which used the principle of Laplacian Pyramid to generate high quality images. However, the images generated by this model were still not photo realistic images but the quality of images was definitely better than the original GANs. The poor quality was because of chaining of multiple models to produce images.

The next GAN model which was introduced made a significant contribution in improving the quality of images generated by a GAN. This GAN is known Direct Convolution GAN (DCGAN).

### A. DCGAN

The DCGAN makes use of the Convolutional Neural Network (CNN) as its Generator and Discriminator [3]. This is one of the first GAN which uses CNN for both the Discriminator and Generator. However, the CNN used in DCGAN is a constrained CNN. Radford et al. [3] suggested that the CNN be constrained architecturally.

One of the main advantages of DCGAN is that it makes stable training of GAN possible. All the GANs before DCGAN were unstable in training which forced the generator to produce impractical images. Due to its stability these GANs are easily scalable. Due to these advantages DCGANs are used in various other applications besides image generation namely: infrared colorization [12], attribute recognition [13], face aging simulation [14].

### B. CGAN

Use either SI (MKS) In an unconditioned General Adversarial Network, no control can be exercised on the mode of the data being generated . In a Conditional GAN, an additional auxiliary input is given to both the generator and the discriminator of the GAN. [15] The introduction of an auxiliary input helps us to gain control and subsequently direct the generation process. This additional input ranges from class labels to data which is obtained from some other modalities. The conditioning is done by feeding this auxiliary input to both the generator and the discriminator. In the generator this additional input is combined with the noise present while at the discriminator with the pre-existing data that is given as input. However, on training the model on unimodal data, results show that the Conditional GAN or CGAN is inferior in performance to several other models, some of which are non- conditional as well. Many application specific frameworks have been modelled on CGAN, where the output has to be obtained from either intricate images of artwork of specific styles (ArtGAN) [16], or face aging (acGAN)[17] to name a few.

## III. TEXT TO IMAGE:

The main problem faced when generating images from text or vice versa is the multimodality in between text and corresponding images. Many images can correspond to the same text input. Because of the adaptive loss function of GANs, they can work around this multimodality problem. Generative Adversarial Text to Image Synthesis [18] by Reed et al proposed a novel model for synthesis of images from detailed textual descriptions. This model is based on DCGAN. GAN - CLS, GAN-INT and GAN-INT-CLS are the three algorithms implemented in this paper. In GAN-CLS, naive GAN Discriminator is modified to detect real/fake image and also to ascertain whether the image matches the text input. In GAN-INT supplementary text embeddings are generated synthetically by interpolating between the dataset embeddings. GAN_INT-CLS is a combination of GAN-INT and GAN-CLS. The images generated by the model mentioned in the paper are of size 64*64 pixels. However, the generated images lack details. Also, generation of images of higher resolution is not possible.

### A. GAWWN

The word "data" is plural, not sin In Learning where and what to draw by Reed et al [19], Generative Adversarial

What-Where Network (GAWWN) is proposed. It improves upon [18] by controlling the location of objects in the synthesized image.

GAWWN implements two models: Bounding box conditional model and Keypoint conditional model.

Bounding box conditional model: Text embeddings are converted to fit in bounded box coordinates. Thus, the location of the objects is controlled by the bounding box coordinates. The object can be scaled, translated and stretched/shrunk when the bounding box coordinates are altered.

Keypoint conditional model: Coordinates of keypoints of the object are used to synthesize the image. By using keypoints to synthesize the image, along with scaling, translating and stretching/shrinking, the orientation of the object can also be changed.

GAWWN has the capability to generate images of size 128*128 pixels. By using keypoint conditional model, important features of image can be detailed. However, this requires providing additional spatial information of the object which can be a drawback.

*B. StackGAN*

It is very difficult to train a GAN to generate images of high resolution. Modifying the GANs mentioned above to generate images of resolution 256*256 pixels leads to training instability. The images synthesized by such GANs are unclear and obscure in nature.

The main advantage of StackGAN is that it can generate photo-realistic images of size 256x256 pixels [6]. The main architectural difference in StackGAN is that two GANs are used together to generate the image. As a result, there are two generators and two discriminators. This network comprising of two GANs helps to overcome the issue of training instability occurring with a single GAN when it is used to generate images of the same resolution.

The text input is passed through an encoder which is a CNN-RNN encoder in this case. Further, Conditioning Augmentation is performed on the encoded text. The pre-processed text is passed to the Stage-I GAN. This GAN generates an image of size 64x64 pixels according to the given text input. This image is further passed on to the Stage-II GAN. The Stage-II takes the image from Stage-I GAN and further enhances its features thereby producing a photo-realistic image. The size of the output image from Stage-II is 256x256 pixel. The block schematic of StackGAN is as mentioned in Fig 1.

The advantage of StackGAN is that they are scalable as bigger images could be constructed by increasing the number of GANs to be trained. However, this increases the computational complexity of the GAN.

*C. Stack GAN ++*

StackGAN ++ by Han Zhang et al [20], generalises StackGAN [6] by arranging multiple generators and discriminators in a format which resembles a tree. Different branches of the tree generate images, from low resolution to high resolution. First branch generates image with preliminary structures and colours while each subsequent branch adds more details. Generators are trained jointly, causing stabilised training of the network. Mode collapse is observed less frequently in StackGAN++ as compared to previously mentioned networks. Thus, the main advantage of this generalized network is improved training stability.

*D. AttnGAN:*

Attentional GAN named AttnGAN [21] by Tao Xu et al. adds attentional mechanism to the GAN framework. It builds upon StackGAN++ [20]. It generates images by combining crucial word level information with overall sentence level information in the text input. AttnGAN is capable of using word level fine grained information corresponding to various subregions of the image.

AttnGAN implements two fundamental components:

Attentional generative network: Along with generating a global sentence vector from the input text, AttnGAN generates multiple word level vectors for conditioning. A basic low resolution image is formed using the global sentence vector while in the following stages the word level vectors are used for generating regional features by using the attentive mechanism. This produces a detailed and high-resolution image in the latter stages.

Deep Attentional Multimodal Similarity Model (DAMSM): It is an additional attentional mechanism for training the generator. The DAMSM ensures that the generated image corresponds to the word level information in the input text as well as agrees with the information contained in the sentence as a whole. By measuring the similarity, it generates a fine-grained loss.

The detailed block schematic of AttnGAN architecture is as mentioned in Fig 2.

AttnGAN is the cumulative result of using an attentive mechanism along with the best features of the GANs mentioned previously. It is currently a state-of-the-art GAN for generation of images from text input and outperforms all the GANs mentioned previously.

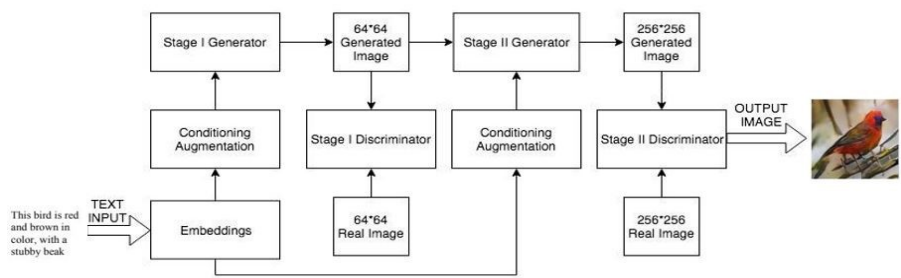

Fig. 1. Block Schematic Of StackGAN

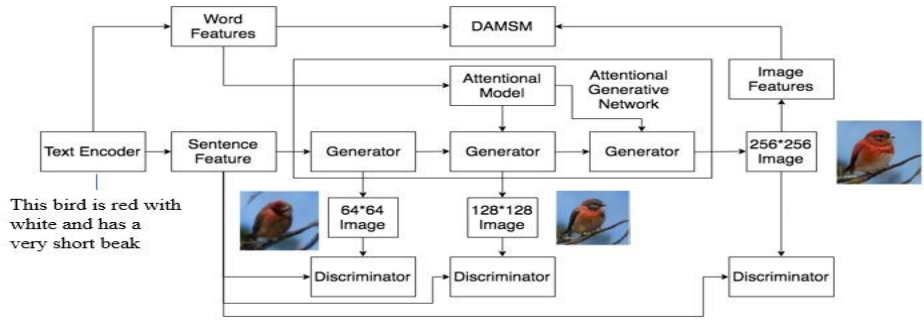

Fig. 2. Block Schematic of AttnGAN

TABLE I. BRIEF SUMMARY OF TEXT TO IMAGE GANs

| Name | Features | Image Resolution | Advantages | Disadvantages |
|---|---|---|---|---|
| GAWWN [19] | • Location constraints are added and are used for conditioning. | 128px×128px | • Controlling the location of object is possible | • Training instability occurs for high resolution image |
| StackGAN [6] | • Two stage GAN are implemented to improve the resolution | 256px×256px | • 256×256 pixel image can be generated<br>• Issue of instability for high resolution images is resolved | • Instability and mode collapse is observed |
| StackGAN++ [20] | • Multiple Generators and Discriminators are arranged in tree format. | 256px×256px | • Generators are trained jointly<br>• Mode collapse is less frequent | • Harder to converge on complex dataset<br>• Requires more GPU memory |
| AttnGAN [21] | • Attentional mechanism is added to StackGAN++ | 256px×256px | • Uses word level fine grained information for conditioning the image.<br>• Significant improvement in quality | • Requires more GPU memory. |

IV. IMAGE TO IMAGE:

These types of GANs take image/images as an input and generate an image as an output. Even though GANs are extremely versatile in nature the main objective of GANs was to generate images from the given input image. Depending on the application of the GAN the output image varies. In this section, we have grouped different GANs according to their application.

A. Cross Domain Relation GANs

Image to image translation from one domain to another is an interesting problem statement which is studied under GANs. Isola et al. [22] converted black and white images to colour images and Taigman et al. [23] generated emojis by using face images. All these solutions relied on paired dataset. However, it is difficult to obtain a paired dataset. To tackle this problem, image to image translation was developed on unpaired datasets. The GANs were responsible for finding their own relations. The state of the art GANs for this application are:

*1) DiscoGAN:*

DiscoGAN refers to Discovering Cross Domain Relations using GAN [7]. DiscoGAN takes an input image from one domain (Domain A) and generates a completely new image belonging to another domain (Domain B). The dataset is unpaired in nature that means that the images in Domain A are not labelled with images in Domain B. This gives much more flexibility to manipulate the data.

DiscoGAN outperforms traditional GAN and GANs with a reconstruction loss as it contains four generators and two discriminators. The first generator takes an input image ($imageA$) from Domain A and generates the image belonging to Domain B ($imageAB$). This $imageAB$ generated acts as an input to the second generator. The second generator reconstructs the image ($imageABA$) belonging to Domain A. The $imageABA$ is compared with the original ($imageA$) belonging to Domain A and the distance is calculated. The $imageAB$ is then passed to discriminator where it is compared with the original image ($imageB$) belonging to Domain B. The same procedure is followed for an image present in Domain B by using the rest of the two generators and a discriminator. This procedure is represented in Fig 3. The reconstruction loss and standard GAN loss is calculated for these generators and discriminators as well. As a result, the DiscoGAN considers the standard GAN loss as well as reconstruction loss during training. This is the reason DiscoGAN outperforms the standard GAN and GANs with reconstruction loss in terms of quality of image generated.

One of the main constraint to DiscoGAN is that every image in Domain A has to be mapped to an image in Domain B. This means that every image in the Domain A has to correspond to one image in Domain B. The mapping from Domain A to Domain B has to be one to one mapping and vice versa else it results in failure of the model. This puts a constraint on selecting the two different domains for generating images.

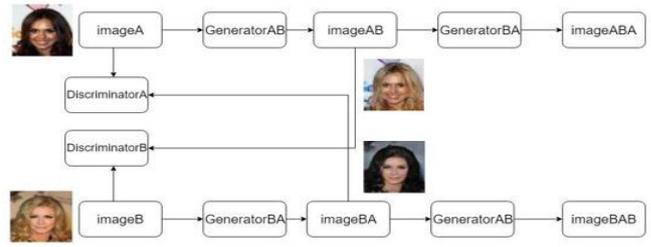

Fig. 3. Block schematic of DiscoGAN

*2) Unpaired image to image translation using Cycle Consistent adversarial networks (CycleGAN)*

A minimum of one CycleGAN learns to map $G: X \rightarrow Y$ such that the distribution of images from G(X) is indistinguishable from the distribution Y using an adversarial loss. [24] As the mapping is highly constrained, CycleGAN couples it with an inverse mapping $F: Y \rightarrow X$ and introduces a cycle consistency loss to enforce F(G(X)) ≈ X. Two adversarial discriminators DX and DY have been introduced in Cycle GAN, where DX aims to distinguish between images {x} and translated images {F(y)}; in the same way, DY aims to discriminate between {y} and {G(x)}, where {$x_i$} represent the training examples. The unpaired image to image translation consists of two auto encoders called as adversarial auto encoders. Each have special internal structure: they map an image to itself via an intermediate representation that is a translation of the image into another domain.

The reason CycleGAN is so efficient is that it implements Cycle Consistent Losses [25] along with adversarial losses.

Even though CycleGAN is one of the most influential GANs used for image to image translation it faces certain issues. The tasks that require geometric changes are not applied with a great precision. For example, on the task of dog to cat transfiguration, the learned translation degenerates to making minimal changes to the input. Similarly, CycleGAN also depends on the dataset used. With a small change in the input image the pretrained CycleGAN finds it tough to find the Cross-Domain relationship. CycleGAN learns to map $G: X \rightarrow Y$ such that the distribution of images from G(X) is indistinguishable from the distribution Y using an adversarial loss. [24] As the mapping is highly constrained, CycleGAN couples it with an inverse mapping $F: Y \rightarrow X$ and introduces a cycle consistency loss to enforce F(G(X)) ≈ X. Two adversarial discriminators DX and DY have been introduced in Cycle GAN, where DX aims to distinguish between images {x} and translated images {F(y)}; in the same way, DY aims to discriminate between {y} and {G(x)}, where {$x_i$} represent the training examples. The unpaired image to image translation consists of two auto encoders called as adversarial autoencoders. Each have special internal structure: they map an image to itself via an intermediate representation that is a translation of the image into another domain.

The reason CycleGAN is so efficient is that it implements Cycle Consistent Losses [25] along with adversarial losses.

Even though CycleGAN is one of the most influential GANs used for image to image translation it faces certain issues. The tasks that require geometric changes are not applied with a great precision. For example, on the task of dog to cat transfiguration, the learned translation degenerates to making minimal changes to the input. Similarly, CycleGAN also depends on the dataset used. With a small change in the input image the pretrained CycleGAN finds it tough to find the Cross-Domain relationship.

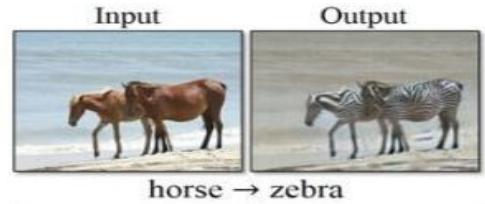

Fig. 4. Block Schematic CycleGAN

B. *Super Resolution*

The main purpose of Super resolution GAN is to up sample the low-resolution image to a high resolution. There have been recent approaches [26] which have discussed the problems of super resolution. The current state of the art GANs are mentioned below.

3.1.1. SRGAN:

SRGAN [27] refers to Super resolution GAN. This GAN generates super resolution images with an upscaling factor as high as 4x. This is the first GAN to do generate a super resolution image with such a high upscaling factor.

To train SRGAN the High Resolution (HR) image is down sampled to Low Resolution (LR) image. This LR image is passed to the Generator. This generator generates a Super Resolution (SR) image which is then passed to the discriminator along with the original HR image to classify the real and fake HR image. The loss calculated is back propagated to train the generator and discriminator.

The main reason SRGAN gives good results is because of its unique loss function. The loss function also known as perceptual loss function is an addition of adversarial loss as well as content loss. Intuitively, the adversarial loss tries to make the image look natural (that is it tries to match the distribution of HR image) and the content loss ensures that the SR image generated has features similar to the lower resolution input image. These two losses combined together make the output image much more realistic copy of the lower resolution counterpart.

The main disadvantage of this architecture is that it is computationally expensive. According to our survey there has not been a substantial study which would produce SR images of quality as high as SRGAN with a smaller architecture. However, the quality of the image obtained from SRGAN is far more superior than any other GAN used for the same purposes.

*C. Sketch to Image:*

The sketch to image GAN takes image as an input and generates an image with colour, texture or better quality pixel value image at its output. One such state-of-the-art GAN is Scribbler that was proposed by Sangkloy et al [28]. The Scribbler converts the sketch images with colour strokes to realistic images. The better version of Scribbler was achieved in Texture GAN as it uses a feed forward network and hence it can run interactively as users modify sketch or texture suggestions. The texture GAN displayed an improvement in precise propagation of colour.

*1) Texture GAN:*

TextureGAN [8], the first deep image synthesis method uses feed-forward network because of which the users can see the after-effect of their edits. With the help of TextureGAN users are able to place two or more patches onto sketch and the network texturizes it to an accurately a 3-D image. Because of the feed forward network it can run interactively as users modify sketch or texture suggestions.

The network is capable of handling multiple texture patches placed on different parts of the images. The network can propagate the textures within semantic regions of the sketch while respecting the sketch boundaries.

TextureGAN makes use of a 5-channel image as input to the network. The channels support three different types of controls – one channel for sketch, two channels for texture (one for intensity and one for binary location mask), and two channels for colour. The detailed block schematic of TextureGAN architecture is as mentioned in Fig 5. This system supports richer user guidance signals including structural sketches, colour patches, and texture swatches. Moreover, it redefines the feature and adversarial losses and introduce new losses to improve the replication of texture details and encourage precise propagation of colours. The style loss, pixel loss and colour loss are used to implement TextureGAN in order to improve the reproduction of texture details, stabilise training for generation of texture details faithful to user input and to introduce colour constraints respectively.

However, TextureGAN cannot reproduce low level texture details since the network focusses on high level structure. Moreover, ablation experiment confirmed that style loss was not enough to encourage texture propagation.

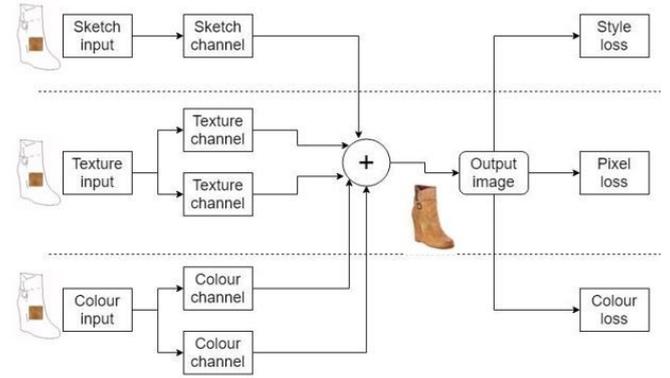

Fig. 5. Loss generation of TextureGAN

*2) SketchyGAN:*

A sketch is generally hand drawn, with certain specific objectives[29]. The GAN takes this sketch as an input image and learns the mapping required to generate the output image, using an added network module knows as Masked Residual Unit within the Generator and Discriminator.

The Masked Residual Unit (MRU) essentially serves as a network module that allows a Convoluted Network to train repeatedly over the same image. The MRU uses a learned internal mask that selects and extract specific features from the given input image and adds them to the features already computed so far. The MRU takes the input feature maps and the given input image as its input values and gives feature maps as its output.

The Generator of sketchy GAN uses an encoder and decoder, both of which are built with MRU blocks. The sketches are resized and fed to every MRU block that lies on the path. To improve the GAN performance, there are skip connections between the encoder and the decoder, so that the output feature blocks from the encoder and the decoder can be concatenated. The discriminator also uses MRU blocks. The detailed block schematic of SketchyGAN architecture is as mentioned in Fig 6.

The main objective behind Sketchy GAN is to generate realism and intent behind the sketch. Unfortunately, the sketchy GAN can only provide one of the two. The images that are generated, though provisionally better than other training models are not as photorealistic as one would expect, and the resolution is not up to the mark.

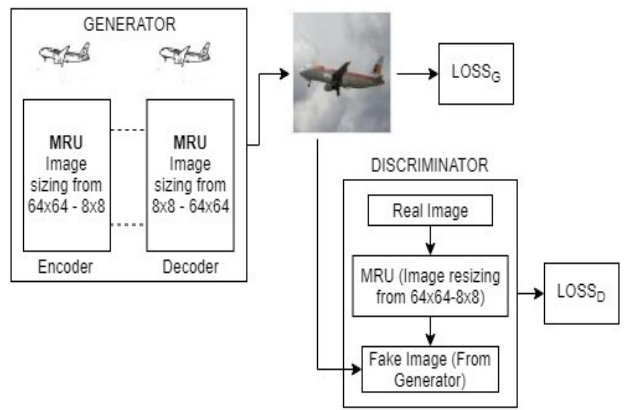

Fig. 6. Block Schematic of SketchyGAN

## D. Special Application GAN

### 1) SemanticGAN:

Semantic GAN [30] proposes a way to semantically manipulate images by text descriptions. Semantic GAN is able to disentangle the semantic information from the two modalities (image and text) and generate new images from the combined semantics.

Semantic GAN architecture basically consists of a generator and a discriminator. The generator is made of an encoder, a residual unit and a decoder. The text description and source image are encoded by the encoder. Semantic GAN adopts the approach of Kiros et al. [31] for pretraining a text encoder that is able to encode text descriptions into visual-semantic text representations. The encoder is generally a Convolutional Neural Network (CNN) while the residual transformation unit is made up of several residual blocks. This makes the generator network easy to learn the identify function, so that the output image would retain similar structure of the source image. it also allows the model to have a deeper encoding process. Finally, images based on both feature representations of images and texts are then synthesized by the decoder.

The SemanticGAN proved a great way to synthesize unseen pictures from the training set. Another advantage of this method was that most of the original background, pose and other information in the original images remain clear.

### 2) ArtGAN:

ArtGAN can be used to generate intricate and abstract images such as artwork [16]. These images may or may not have definite backgrounds, foregrounds or may consist of more than one object per image. This is achieved by training the GAN to focus on a specific subject by giving some additional input to it. We feed an additional vector "y" as an additional input layer into the discriminator and generator. Moreover, the errors generated in the output are back propagated to the generator as well.

The main feature in the structure of ArtGAN is that there is a feedback from the labels, or additional input labels of each generated image through the loss function to the generator. Moreover, layer "y" is given by the user and encodes the information of either the attributes or the classes of data to control the modes of data to be generated.

These modifications to the General Adversarial Networks architecture enable the ArtGAN to learn and generate complex and abstract images, of varying artistic styles.

TABLE II. BRIEF SUMMARY OF DIFFERENT GANS USED USED FOR IMAGE GENERATION

| Application | GAN Name | Features | Disadvantage |
|---|---|---|---|
| Super Resolution | SRGAN [27] | <ul><li>It can be used to upsample images to a 4x resolution</li><li>State of the art for generating great quality image for 4x upscaling resolution</li><li>Best GAN available for upscaling the images</li></ul> | <ul><li>Computationally expensive</li><li>Large architecture</li></ul> |
| Cross Domain Relation GAN | DiscoGAN [7] | <ul><li>DiscoGAN produces images on unpaired datasets which reduces the problems of datasets avalaibility.</li><li>Due to large number of Generators and Discriminators and unique loss function, DiscoGAN produces Cross Domain images which are much more superior that its predecessors</li></ul> | <ul><li>Every image in first domain has to be mapped to another image in second domain. That is one to one mapping is a necessity in DiscoGAN which puts a constraint on the selected dataset.</li></ul> |
| Cross Domain Relation GAN | CycleGAN [24] | <ul><li>CycleGAN produces images on unpaired datasets which reduces the problem of data availability</li><li>It implements Cycle Consistent Losses along with adversarial loss which improves the efficiency of the of image to image translation quality.</li></ul> | <ul><li>The tasks that require geometric changes are not applied with a great precision in CycleGAN.</li></ul> |
| Sketch to Image GAN | TextureGAN [8] | <ul><li>The objective function consists of content loss, feature loss, texture loss and adversarial loss which improves the quality of image</li><li>It consists of feed-forward network which allows to run interactively as users modify sketch or texture suggestion</li><li>Capable of handling multiple texture patches</li></ul> | <ul><li>Difficulty for the network in reproducing low level texture details as the network focuses on high level structure, colour, pattern.</li><li>The texture elements in the camera-facing centre were found to be larger than those around the boundary.</li><li>Textures at the bottom of the objects were often shaded darker than the rest.</li></ul> |
| | SketchyGAN [29] | • Uses Masked Residual Unit (MRU) that trains a convolutional network repeatedly on the same input image to extract newer features and coalesce them with the features already extracted<br>• Performs better than other existing sketch to image transformation networks. | • Provides poor resolution of the output image.<br>• Results are not photorealistic, fails to capture human intent of the sketch due to lack of training pairs of sketch and photos |
| Special Application GAN | ArtGAN [16] | • Uses cross entropy to back propagate errors to Generator through feedback network. This helps the generator train faster and the framework can grasp abstract objects and artistic styles better. | |
| | SemanticGAN [30] | • Semantic GAN uses a specific loss function called the adaptive loss for semantic image synthesis.<br>• Most of the original background, pose and other information in the original images remain clear. | • Due to the limited size of dataset employed, its encoder may not be capable of producing good representations. |

## V. CONCLUSION AND FUTURE WORK

Generative Adversarial Networks have pioneered unsupervised and semi-supervised deep learning methods. The profusion of interest and innovation in GANs can be credited to their versatility in applying themselves in image, text and music generation of various kinds. While Text to image conversion frameworks such as StackGAN can generate photorealistic images of 256x256 pixels, image to image conversion frameworks such as DiscoGAN are able to work with unpaired datasets, which mitigates the problem of dataset availability. Frameworks such as SRGAN generate sharper high-resolution images of upto 4x upscaling and can also be used to generate images from sketches or intricate artwork and are shown to have promising results.

Inspite of showing better results than most other frameworks, GANs struggle in certain aspects. Firstly, they are unable to generate images of a resolution higher than 256x256 pixels. This is due to the high computational cost incurred when dealing with images of a higher resolution While GAN frameworks can be used for video synthesis of 3D models, the results obtained are far from optimal. Apart from this mode collapse, high sensitivity to hyperparameters and a diminishing gradient in the case of an overly successful discriminator are problems that plague GAN frameworks. While improvements are being made to resolve these issues and attain better results, they require more exploration and research to provide tangible results.

As the potential and applicability of GAN's are immense, techniques which overcome the issues and limitations mentioned above need to be researched in the future.